\documentclass[letterpaper, 10 pt, conference]{ieeeconf} 
\IEEEoverridecommandlockouts
\usepackage{amsmath}
\usepackage{amssymb}

\usepackage{graphicx}
\usepackage{capt-of}
\usepackage{overpic}
\usepackage[export]{adjustbox}
\usepackage{minibox}
\usepackage{textcomp}
\usepackage{tabularx}
\usepackage{multirow}
\usepackage{booktabs}
\usepackage{arydshln}
\usepackage{adjustbox}
\usepackage{pgfplots}
\usepackage{siunitx}
\usepackage[implicit=false]{hyperref}
\usepackage{pifont}
\usepackage{stfloats}
\usepackage{makecell}

\usepackage{algpseudocode,algorithm,algorithmicx}

\algrenewcommand\algorithmicrequire{\textbf{Precondition:}}
\algrenewcommand\algorithmicensure{\textbf{Postcondition:}}
\newcommand{\cmark}{\ding{51}}%
\newcommand{\xmark}{\ding{55}}%

\usepackage[caption=false, font=scriptsize]{subfig}

\title{\LARGE \bf
Semantic Map Learning of Traffic Light to Lane Assignment\\
based on Motion Data
}

\author{Thomas Monninger$^{1}$, Andreas Weber$^{2}$, Steffen Staab$^{3,4}$%
\thanks{$^{1}$Mercedes-Benz Research \& Development North America, Sunnyvale, CA, USA (email: thomas.monninger@mercedes-benz.com)}%
\thanks{$^{2}$Mercedes-Benz AG, Research \& Development, Stuttgart, Germany (email: andreas\_silvius.weber@mercedes-benz.com)}%
\thanks{$^{3}$University of Stuttgart, Institute of Parallel and Distributed Systems, Stuttgart, Germany (email: steffen.staab@ipvs.uni-stuttgart.de)}%
\thanks{$^{4}$University of Southampton, Electronics and Computer Science, Southampton, United Kingdom}%
\thanks{$^{5}$\url{https://github.com/map-learning/tl2la}}%
}

\newcommand\copyrighttext{\footnotesize \textcopyright~2023 IEEE. Personal use of this material is permitted.  Permission from IEEE must be obtained for all other uses, in any current or future media, including reprinting/republishing this material for advertising or promotional purposes, creating new collective works, for resale or redistribution to servers or lists, or reuse of any copyrighted component of this work in other works.
}%

\newcommand\copyrightnotice{%
	\begin{tikzpicture}[remember picture,overlay]
		\node[anchor=south,xshift=0pt,yshift=2pt] at (current page.south) {\fbox{\parbox{\dimexpr\textwidth-\fboxsep-\fboxrule\relax}{\copyrighttext}}};
	\end{tikzpicture}%
}

\begin{document}
\maketitle
\thispagestyle{empty}
\pagestyle{empty}

\begin{abstract}
Understanding which traffic light controls which lane is crucial to navigate intersections safely.
Autonomous vehicles commonly rely on High Definition (HD) maps that contain information about the assignment of traffic lights to lanes.
The manual provisioning of this information is tedious, expensive, and not scalable.
To remedy these issues, our novel approach derives the assignments from traffic light states and the corresponding motion patterns of vehicle traffic.
This works in an automated way and independently of the geometric arrangement.
We show the effectiveness of basic statistical approaches for this task by implementing and evaluating a pattern-based contribution method.
In addition, our novel rejection method includes accompanying safety considerations by leveraging statistical hypothesis testing.
Finally, we propose a dataset transformation to re-purpose available motion prediction datasets for semantic map learning.
Our publicly available API for the Lyft Level 5 dataset enables researchers to develop and evaluate their own approaches$^5$.
\end{abstract}

\section{Introduction}
\copyrightnotice
Autonomous vehicles require a semantic understanding of the given traffic scene to navigate complex environments safely.
At intersections, understanding the assignment of traffic lights to lanes is a prerequisite to determining whether to stop.
This assignment information is used in safety-critical applications and currently cannot be derived by an online system with the required reliability.

The traffic light to lane assignment (TL2LA) is defined by the geometric arrangement of traffic lights relative to the lanes in an intersection and optionally by indication inlays inside the traffic light bulbs, such as arrows.
Using this information to automate the map annotation in a scalable way does not reach the required level of correctness.
The vast variety of geometric configurations of traffic lights and lanes and the unreliable detection of traffic light inlays make it very challenging to precisely assign individual traffic lights to their respective lanes.
Hence, the assignment is traditionally provided a priori from an HD map, which involves laborious manual annotation efforts.
Modeling the broad varieties of intersection branches, topology, arrangement of traffic lights, etc., constitutes its own modeling and knowledge acquisition problem \cite{Ulbrich2014}.
Fully capturing the geometric layout and semantic relationships between traffic elements of an urban traffic scene may be difficult even for humans.

Current research has rarely considered the learning of semantic map information.
One reason is that systems with high level of autonomy (e.g., robotaxis) use an HD map as a strong prior to meet the essential requirement for high safety.
On a limited scale, HD maps can be precisely annotated by humans, dropping the need for automated solutions.
Secondly, the limited prior work has focused on deriving TL2LA from the geometric arrangement of traffic lights and lanes at intersections.
These approaches require labeled data for a multitude of intersection layouts to generalize well.

\begin{figure}[!t]
	\vspace{2mm}
	\includegraphics[width=1\columnwidth,trim={0 0 0 0.0cm},clip]{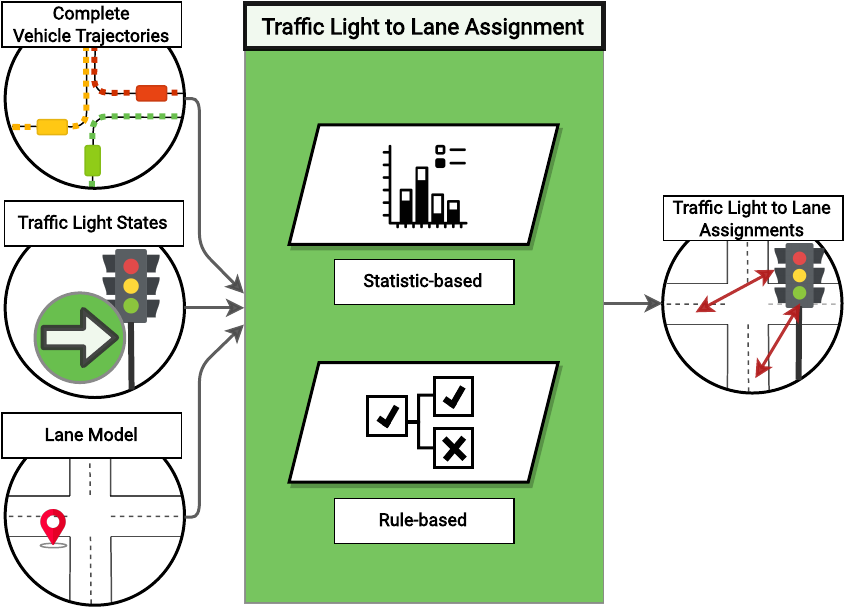}
	\caption{Overview of our approach to derive traffic light to lane assignments from detected traffic light states and motion patterns of vehicle traffic.}
	\label{fig:overview}
	\vspace{-2mm}
\end{figure}

In this paper, we propose a novel paradigm of learning semantic map features from motion patterns of vehicle traffic.
Our main observation is that humans resolve the TL2LA problem while driving, leading to motion patterns that implicitly contain the required data.
We implement two methods for statistical and rule-based discovery of the TL2LA from detected traffic light states and the respective motion patterns as shown in Figure \ref{fig:overview}.
This approach is independent of the geometric arrangement and therefore generalizes to any country and intersection layout.
Finally, we present a dataset transformation to re-purpose available datasets for this task and evaluate the two proposed methods.

In summary, our main contributions are:
\begin{itemize}
	\item We propose a novel learning task, deriving the TL2LA from motion patterns of traffic, and provide two methods to solve this task including safety considerations.
	\item We present a dataset transformation that changes the representation of motion prediction datasets to make them suitable for learning map semantics.
	\item We evaluate our methods on the transformed Lyft Level 5 dataset and provide an API for future research in the field of motion-based semantic map learning.
\end{itemize}

This work is structured as follows:
Section \ref{sec:related_work} discusses related work.
Section \ref{sec:approach} presents problem domain, problem statement and our approach.
Implementation and experiments are described in sections \ref{sec:implementation} and \ref{sec:experiments}.
Finally, section \ref{sec:limitations} covers remaining limitations and section \ref{sec:conclusion} gives a conclusion.

\begin{table*}[b]
	\centering
	\caption{Comparison of the dataset meta information provided by different motion prediction datasets.}
	\label{tab:comparison_metadata}
	\resizebox{\textwidth}{!}{
	\begin{tabular}{lccccl}
		\toprule
		\textbf{Meta Information}                                                  & \textbf{Argoverse 2.0 \cite{Argoverse2}} & \textbf{nuScenes \cite{Nuscenes2020}} & \textbf{Waymo Motion \cite{WaymoMotionDataset2021}} & \textbf{Lyft Level 5 \cite{WovenPlanet2020}} & \textbf{Requirements Map Learning}                                                                     \\
		\midrule
		Number of scenes~$N$                                                           & \SI{250000}{}                & \SI{41000}{}           & \SI{103000}{}               & \SI{162000}{}            & As high as possible           \\
		
		Scene duration~$\tau$                                                         & \SI{11}{\s}                    & \SI{20}{\s}              & \SI{20}{\s}                    & \SI{25}{\s}                & At least multiple seconds                                                                                    \\
		
		Unique roadways~$R$                                                            & \SI{2110}{\km}                 & \SI{4300}{\km}           & \SI{1750}{\km}                & \SI{10}{\km}               & High as long as density high                                                                                       \\
		
		Spatial recording density~$\rho$ & \SI{1303}{\s\per\km}                   & \SI{190}{\s\per\km}             & \SI{529}{\s\per\km}                  & \SI{405000}{\s\per\km}            & As high as possible                                                                                    \\
		
		Total time                                                             & \SI{763}{\hour}                  & \SI{228}{\hour}             & \SI{570}{\hour}                  & \SI{1125}{\hour}              & As high as possible                                                                                    \\
		
		Size                                                                       & \SI{58}{GB}                    & \SI{48}{GB}             & \SI{1.4}{TB}                 & \SI{78}{GB}               &                                                                   \\
		
		Coordinate System                                                          & global                 & global           & local                 & global             & Global to localize scenes \\
		
		License                                                                    & CC BY-NC-SA 4.0        & CC BY-NC-SA 4.0  & CC BY-NC-SA 4.0       & CC BY-NC-SA 4.0    &                                                                               \\
		
		\bottomrule
	\end{tabular}}
\end{table*}

\section{Related Work}\label{sec:related_work}
This section reviews related work concerning semantic map learning, deriving information from motion data and available datasets.

\subsection{Semantic Map Learning} \label{sec:semantic_map_learning}
Maps comprise geometric objects and semantic features, which are both needed to perform the driving task.
Geometric objects, such as signs and crosswalks, are directly perceivable by sensors.
Semantic map features are virtual and relate to geometric objects in the real world.
Those include derived entities such as lanes, which model the lane corridor and are constrained by the geometric lane dividers.
Another type of semantic map features are relationships between map entities, i.e., the TL2LA.

One way of deriving the TL2LA is by applying heuristics to the geometric arrangement.
Early work from Fairfield and Urmson \cite{Fairfield2011} uses a simple heuristic to add the TL2LA to a geometric map of the traffic lights.
They acknowledge the challenge of this task and integrate a human verification step in their process because the heuristic alone did not provide the desired quality.
Poggenhans \cite{Poggenhans2019} defines a more methodical approach by developing a rule set from the official traffic regulations as a heuristic.
Similarly, that approach derives a geometric arrangement of the road infrastructure first.
In a second step, the proposed rule set is used to solve semantic relations such as right of way and TL2LA.
Again, the concluding statement indicates that heuristic approaches are insufficient because they rely on correct and unambiguous infrastructure, which is not always given.

Alternative ways of deriving TL2LA include learning-based approaches on the geometric arrangement.
Li et al. \cite{Li2021} formulate the problem of semantic map learning and predict semantic map geometries in bird's-eye view from sensor data.
Similar to other works \cite{Zhou2021, bevformer, Elhousni2020}, they fall short of deriving semantic relations like the TL2LA.
Langenberg et al. \cite{Langenberg2018} target the TL2LA in image space.
They transform the lower part of the camera image using an inverse perspective mapping to get a top-down view of the road surface and use a learning-based model to predict the TL2LA directly from that image.

All approaches listed above tackle the problem of semantic map learning by deriving semantics from the geometric representation.
This requires a diverse dataset to cover all variations of the real world.
In contrast, our approach does not use geometry information of traffic lights, i.e., their position and heading are not input to the approach.

\subsection{Deriving Information from Motion Data}
Prior work has used motion data in the form of vehicle trajectories to derive information.

There is substantial literature on general clustering and pattern recognition on motion data.
Yuan et al. \cite{Yuan2017} review moving object trajectory clustering algorithms.
While they do not target semantic map learning, the methods described are a precondition of deriving information from massive motion data.
Jiao et al. \cite{Jiao2004} show an approach for characterizing motion data based on summarizing and classifying patterns.
These works do not attempt TL2LA but provide general ideas for working with motion data.

Scientific work in the context of mapping from motion data mostly targets geometric map features.
Early work from Chen and Krumm \cite{Chen2010} and from Uduwaragoda et al. \cite{Uduwaragoda2013} use statistical models to predict traffic lanes from GPS traces.
More recent work aims to close the gap to HD maps by deriving additional geometric map features such as boundaries and signs \cite{Doer2020} by using higher-level features such as lane marking types provided by the vehicle fleet \cite{Liebner2019} or by directly extracting lane-level information from raw motion patterns \cite{Shu2020}.

Only a few publications describe how to derive semantic map attributes from motion data.
Derrow-Pinion et al. \cite{Derrow2021} predict the estimated time of arrival for a queried map route using motion data transmitted from mobile devices.
Wirthmüller et al. \cite{wirthmuller2021} model a semantic lane attribute, the probability of lane change events, based on data from the vehicle fleet.
Our novel approach aims to derive the TL2LA purely from motion data, a topic that has not yet been explored in the literature.

\subsection{Available Datasets}
High-quality, large-scale datasets are crucial to train models for autonomous driving, especially in the mapping domain with extremely diverse real-world scenarios and highly manual labeling efforts.
Unfortunately, there are no publicly available datasets that specifically address the problem of semantic map learning. 
However, there are several public motion prediction datasets that address the primary goal of motion forecasting in urban environments \cite{Argoverse2, Nuscenes2020, math8081368, WaymoMotionDataset2021, WovenPlanet2020}.
They contain time sequences of motion data, and some additionally include HD maps with semantic annotations for improved motion prediction. 

Table \ref{tab:comparison_metadata} lists publicly available motion prediction datasets and gives an overview of their metadata.
The rightmost column addresses requirements for semantic map learning, which are discussed in our approach.

\section{Motion-based Approach to derive TL2LA} \label{sec:approach}
This section formalizes the problem domain and problem statement of deriving the TL2LA based on motion patterns. 
We propose two methods to solve the problem and explain safety considerations.

\subsection{Formalizing the Problem Domain} \label{sec:problem_domain}
Let $V$ be the set of all vehicles and $\operatorname{pos}: V \times T \rightarrow \mathbb{R}^{2}$ be a partial function that maps each vehicle to its position at time~$t \in T$.
Kinematics are given by the functions $\operatorname{vel}(v,t) = \partial{\operatorname{pos}(v,t)} / \partial{t}$ for velocity and $\operatorname{acc}(v,t) = \partial{\operatorname{vel}(v,t)} / \partial{t}$ for acceleration of vehicle $v$ at time $t$.
Let $S$ be the set of all traffic lights and $\operatorname{state}: S \times T \rightarrow \{\text{red}, \text{green}\}$ be a partial function that maps each traffic light to its state (red or green) at time~$t \in T$.
Let $L$ be the set of all lanes on the road and $\operatorname{boundary}(l) = (b_l, b_r) \; \text{with} \; b \in \mathbb{R}^{2 k} \times \mathbb{R}^{2 k}$ be a function that maps each lane to its left and right boundary~$b$ as a sequence of $k$ points in $\mathbb{R}^2$.
The TL2LAs are represented by a function $\operatorname{assign}: S \times L \rightarrow \{0,1\}$, which maps each pair of traffic light and lane to its binary value (0 for no assignment, 1 for assignment).

\subsection{Problem Statement} \label{sec:problem}
The required input data for deriving TL2LA from motion data comprises the lane geometry ($\operatorname{boundary}$), the position of vehicles over time ($\operatorname{pos}$), and the traffic light state over time ($\operatorname{state}$).
Lane topology and the mapping of vehicles to lanes are optional, because these can be derived geometrically.

We assume to have a dataset:

\begin{equation}
	D = \left( V, S, L, \operatorname{pos}, \operatorname{state}, \operatorname{boundary} \right)
\end{equation}

Predicting the TL2LA is a binary classification problem.
Let target output~$Y$ be the TL2LA for each pair of traffic light and lane, as represented by the definition of the function $\operatorname{assign}$:

\begin{equation}
	Y = \operatorname{assign}(s,l)
\end{equation}

Then the goal is to find a function:
\begin{equation}
	\label{eq:tl2la_dataset}
	g \left( V, s, l, \operatorname{pos}, \operatorname{state}, \operatorname{boundary} \right) = Y = \operatorname{assign}(s,l)
\end{equation}

\subsection{Methods to derive TL2LA}\label{sec:methods}
\begin{figure}[tpb]
	\vspace{2.3mm}
	\centering
	\includegraphics[width=.31\textwidth,trim={0 2 0 0.0cm},clip]{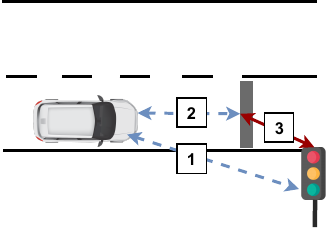}
	\caption{Steps to derive TL2LA (3) based on the traffic light to vehicle (1) and vehicle to lane assignment (2).}
	\label{fig:tl2la}
	\vspace{-2.3mm}
\end{figure}

To get from temporal correspondences between motion patterns of vehicles and traffic light states to TL2LAs, the following procedure is needed as shown in Figure \ref{fig:tl2la}:
\begin{enumerate}
	\item Based on motion pattern $\operatorname{pos}(v,t) \operatorname{for} t \in T$ of vehicle~$v$ and the state over time $\operatorname{state}(s,t) \operatorname{for} t \in T$ of traffic light~$s$: Derive evidence for or against an assignment between $v$ and $s$.
	\item Based on geometric position $\operatorname{pos}(v,t)$ and lane geometries $\operatorname{boundary}(l) \operatorname{for} l \in L$: Use geometric matching $\operatorname{loc}: V \times T \rightarrow L$ to assign vehicle $v$ to its driving lane $l$ at time $t$.
	\item Determine the TL2LA $\operatorname{assign}(s,l)$ between traffic light~$s$ and lane~$l$ by aggregating individual evidences predicting the relation between $l$ and $s$.
\end{enumerate}

We propose two statistical methods and compare with a naive baseline.

\subsubsection{Naive Baseline Method}
Predict the class with the highest prior probability, which can be formalized as:
\begin{equation}
	\operatorname{assign}_{\operatorname{prior}}(s,l) = \operatorname*{arg\,max}_{c_i \in C} \, n_i 
\end{equation}
where $C = $ \{0, 1\} is the set of TL2LA classes (0 for no assignment, 1 for assignment) and $n_i$ is the number of occurrences of class $c_i$ in the dataset.

\subsubsection{Pattern-based Contribution Method}
This method extracts motion patterns individually from each point in time and weighs the contributions w.r.t. the TL2LAs.
A heuristic function $h: V \times S \times T \times \operatorname{Cond} \rightarrow \mathbb{R}$ calculates a contribution value given a vehicle, a traffic light state, a point in time and $\operatorname{Cond}$, a set of condition functions defined below.
The heuristic function $h$ is defined such that positive values contribute to an assignment and negative values have the opposite effect.
The decision about an assignment is made by summing up all individual contribution values that were generated by the heuristic function:
\begin{equation}
	\operatorname{assign}_{\operatorname{pattern}}(s,l) = \sum_{t \in T} \sum_{(v,s) \in V \times S} h \left( v, s, t, \operatorname{cond} \right)
\end{equation}

\begin{table*}[tpb]
    \vspace{2mm}
    \centering
    \caption{Heuristic function $h$ based on the velocity ($\operatorname{vel}$) and acceleration ($\operatorname{acc}$) of vehicle and traffic light (TL) state.}
    \label{tab:heuristic}

    { \def\arraystretch{0.3}
        \begin{tabular}{%
            >{\raggedright}p{.1\textwidth}%
            >{\raggedright}p{.1\textwidth}%
            >{\centering\arraybackslash}p{.09\textwidth}%
            >{\raggedright}p{.50\textwidth}%
            >{\raggedleft\arraybackslash}p{.09\textwidth}%
            }
            \toprule
            \textbf{Pattern}                                                                                & \textbf{Kinematics}                                                                                  & \textbf{TL State}             & \textbf{Additional Conditions}                                                                                                                                                                                                       & \multicolumn{1}{l}{\textbf{Heuristic} $h$} \\
            \midrule
            \multirow{3}{*}[-0.8em]{Stationary}                                                             & \multirow{3}{*}[-0.4em]{\begin{tabular}[c]{@{}l@{}} $ \begin{aligned} \lvert \operatorname{vel} \rvert &< 1 \; \wedge\\[-0.25em] \lvert \operatorname{acc} \rvert &< 1 \end{aligned}$ \end{tabular}}      & red                           & $(\operatorname{distance} < \operatorname{stop\_zone})$                                                                                                                                                                              & $+2 \qquad$                                \\
            \cmidrule{3-5}
                                                                                                            &                                                                                                      & \multirow{2}{*}{green}        & $\operatorname{is\_lead}$                                                                                                                                                                                                            & $-1 \qquad$                                \\
            \cmidrule{4-5}
                                                                                                            &                                                                                                      &                               & $\operatorname{is\_lead} \; \wedge \;  (\operatorname{state\_duration} > \operatorname{reaction\_time\_green})$                                                                                                                      & $-3 \qquad$                                \\
            \midrule
            \multirow{4}{*}[-1.5em]{\begin{tabular}[c]{@{}l@{}}Continuously \\moving\end{tabular}}          & \multirow{5}{*}[-1.2em]{\begin{tabular}[c]{@{}l@{}} $ \begin{aligned}\lvert \operatorname{vel} \rvert &\ge 1 \; \wedge\\[-0.25em] \lvert \operatorname{acc} \rvert &< 1 \end{aligned}$ \end{tabular}}      & \multirow{2}{*}{red}          & $(\operatorname{distance} < \operatorname{slow\_zone})$                                                                                                                                                                              & $-1 \qquad$                                \\
            \cmidrule{4-5}
                                                                                                            &                                                                                                      &                               & \begin{tabular}[c]{@{}l@{}}$(\operatorname{distance} < \operatorname{stop\_zone}) \; \wedge \; (\operatorname{state\_duration} > \operatorname{reaction\_time\_red})$  \end{tabular}                                                                             & $-3 \qquad$                                \\
            \cmidrule{3-5}
                                                                                                            &                                                                                                      & \multirow{2}{*}{green}        & $(\operatorname{distance} < \operatorname{slow\_zone})$                                                                                                                                                                              & $+3 \qquad$                                \\
            \cmidrule{4-5}
                                                                                                            &                                                                                                      &                               & $(\operatorname{distance} < \operatorname{stop\_zone})$                                                                                                                                                                              & $+5 \qquad$                                \\
            \midrule
            \multirow{3}{*}[-0.3em]{\begin{tabular}[c]{@{}l@{}}Acceleration \\from stationary\end{tabular}} & \multirow{4}{*}[0.5em]{\begin{tabular}[c]{@{}l@{}l@{}}$ \begin{aligned} \lvert \operatorname{vel} \rvert &< 1 \; \wedge \\[-0.25em]  \lvert \operatorname{acc} \rvert &\ge 1 \; \wedge\\[-0.25em] \operatorname{acc} &>  0 \end{aligned}$ \end{tabular}}    & \multirow{2}{*}[0.5em]{red}   & \begin{tabular}[c]{@{}l@{}}$(\operatorname{distance} < \operatorname{stop\_zone}) \; \wedge \; (\operatorname{turn\_type} \ne \operatorname{right\_turn})$  \end{tabular}                                                                       & $-2 \qquad$                                \\
            \cmidrule{3-5}
                                                                                                            &                                                                                                      & \multirow{2}{*}[0.5em]{green} & \begin{tabular}[c]{@{}l@{}}$(\operatorname{distance} < \operatorname{slow\_zone}) \; \wedge$ \\ $(\operatorname{state\_duration} > \operatorname{reaction\_time\_red})$  \end{tabular}                                                                       & $+3 \qquad$                                \\
            \midrule
            \multirow{3}{*}[-0.7em]{\begin{tabular}[c]{@{}l@{}}Acceleration \\while moving\end{tabular}}    & \multirow{4}{*}[+0.2em]{\begin{tabular}[c]{@{}l@{}l@{}}$ \begin{aligned} \lvert \operatorname{vel} \rvert  &\ge 1 \; \wedge \\[-0.25em]  \lvert \operatorname{acc} \rvert &\ge 1 \; \wedge\\[-0.25em] \operatorname{acc} &> 0 \end{aligned}$ \end{tabular} } & \multirow{2}{*}{red}          & $(\operatorname{distance} < \operatorname{slow\_zone})$                                                                                                                                                                              & $-1 \qquad$                                \\
            \cmidrule{4-5}
                                                                                                            &                                                                                                      &                               & \begin{tabular}[c]{@{}l@{}}$(\operatorname{distance} < \operatorname{stop\_zone}) \; \wedge \; (\operatorname{state\_duration} > \operatorname{reaction\_time\_red})$    \end{tabular}                                                                       & $-3 \qquad$                                \\
            \cmidrule{3-5}
                                                                                                            &                                                                                                      & green                         & $(\operatorname{distance} < \operatorname{slow\_zone})$                                                                                                                                                                              & $+1 \qquad$                                \\
            \midrule
            \multirow{3}{*}[-1.1em]{Deceleration}                                                           & \multirow{5}{*}[+0.2em]{\begin{tabular}[c]{@{}l@{}l@{}}$\begin{aligned} \lvert \operatorname{vel} \rvert &\ge 1 \; \wedge \\[-0.25em]  \lvert \operatorname{acc} \rvert &\ge 1 \; \wedge\\[-0.25em] \operatorname{acc}  &< 0 \end{aligned}$\end{tabular} }   & red                           & $(\operatorname{distance} < \operatorname{slow\_zone})$                                                                                                                                                                              & $+2 \qquad$                                \\
            \cmidrule{3-5}
                                                                                                            &                                                                                                      & \multirow{2}{*}{green}        & \begin{tabular}[c]{@{}l@{}}$(\operatorname{distance} < \operatorname{stop\_zone}) \; \wedge \; \operatorname{is\_lead} \; \wedge \; (\operatorname{turn\_type} = \operatorname{left\_turn} )$\end{tabular}    & $-1 \qquad$                                \\
            \cmidrule{4-5}
                                                                                                            &                                                                                                      &                               & \begin{tabular}[c]{@{}l@{}}$(\operatorname{distance} < \operatorname{stop\_zone}) \; \wedge \; \operatorname{is\_lead} \; \wedge \; (\operatorname{turn\_type} =  \operatorname{straight} )$\end{tabular} & $-2 \qquad$                                \\
            \midrule
            Other                                                                                           & {~~~$\operatorname{else}$}                                                                           & red/green                     &                                                                                                                                                                                                                                      & $ 0 \qquad$                                \\
            \bottomrule
        \end{tabular}
    }
\end{table*}
Table \ref{tab:heuristic} shows the definition of $h$ with the set of condition functions $\operatorname{cond} = \{\operatorname{vel}$, $\operatorname{acc}$, $\operatorname{distance}$, $\operatorname{is\_lead}$, $\operatorname{turn\_type}$, $\operatorname{state\_duration}\}$.
Those are a)~the distance of a vehicle to the intersection $\operatorname{distance}: V \times T \rightarrow \mathbb{R}$ with two thresholds $\operatorname{stop\_zone}=\SI{8}{\m}$ and $\operatorname{slow\_zone}~=~\SI{20}{\m}$,
b)~whether the vehicle is the first vehicle before the intersection entry $\operatorname{is\_lead}: V \times T \rightarrow \{\operatorname{true}, \operatorname{false}\}$,
c)~the turn information of the current lane $\operatorname{turn\_type}: L \rightarrow \{\operatorname{left\_turn}, \operatorname{right\_turn}, \operatorname{straight}\}$,
d)~the duration of the current traffic light state $\operatorname{state\_duration}: S \times T \rightarrow \mathbb{R}$ relative to the time delay of a vehicle reacting to a traffic light state change ($\operatorname{reaction\_time\_red}=\SI{1}{\s}$ and $\operatorname{reaction\_time\_green}=\SI{3}{\s}$).
By applying this set of rules, contribution values for a pair of traffic light and lane are calculated for each point in time.
When the aggregated value of all contributions for a pair of traffic light and lane is positive, the method predicts $\operatorname{assign}(s,l) = 1$, otherwise it predicts $\operatorname{assign}(s,l) = 0$.
Predictions are aggregated over all points in time based on the majority class.

\subsubsection{Rejection Method}
The rejection method formulates the problem as a hypothesis test.
Conservatively assume an assignment for all pairs of traffic lights and lanes ($H_0$): $\forall (s,l) \in S \times L: \operatorname{assign}(s,l) = 1$.
Set $\operatorname{assign}(s,l) = 0$ only if a significant number of vehicles have been recorded passing the intersection on lane $l$ while $\operatorname{state}(s,t)=\text{red}$.
$$\begin{array}{ll}
	H_0: & \operatorname{assign}(s,l) = 1 \quad \text{(assignment)}
	\\
	H_1: & \operatorname{assign}(s,l) = 0 \quad \text{(no assignment)}
\end{array}$$

The detection of a vehicle passing the intersection is defined relative to the intersection entry.
If the velocity of a vehicle directly in front of the intersection entry (${\mathrm{distance} < \SI{1}{\meter}}$) is greater than a specific threshold (${\operatorname{vel} > \SI{15}{\km\per\hour}}$), a pass is assumed.

A binomial hypothesis test is used to reject the null hypothesis based on a significance level.
By initially assuming a true assignment, this method minimizes false negatives and optimizes for recall.
A further advantage is that this method provides an output for all pairs of traffic lights and lanes.
We assume that a TL2LA exists if less than 5 \% of recorded vehicles pass on a red light (binomial distribution with $p=0.05$).
To minimize the likelihood of false rejections of $H_0$, we choose a significance level $\alpha = 0.001$.
For a pair of traffic light $s$ and lane $l$, let $n$ be the overall number of passes in the dataset where $\operatorname{loc}(v,t) = l \; \operatorname{for} \; v \in V, t \in T$ while the traffic light $s$ is detected with $\operatorname{state}(s,t)$ and let $k$ be the number of the subset of passes with $\operatorname{state} = {\operatorname{red}}$.
Then the TL2LA is derived as:

\begin{equation}
	\operatorname{assign}_{\operatorname{rejection}}(s,l) = \left[ \operatorname{Binomialtest}(k,n,p) < \alpha \right]
\end{equation}

The concept of right turn on red contradicts the traffic light to lane assignment, since vehicles are allowed to pass the red light after stopping for crossing traffic.
As a special heuristic, our rejection method only invalidates the assignment of a traffic light to a right-turning lane for passes with a significantly higher velocity ($v > \SI{25}{\km\per\hour})$.
Unprotected turns are handled implicitly by the rejection method, since it does not extract evidence from stopping vehicles, but purely invalidates TL2LAs given passes on red light.

\subsection{Safety Considerations} \label{sec:safety}
To ensure maximum safety, an autonomous vehicle should not pass an intersection entry when any of the assigned traffic lights is red.
Therefore, false negative TL2LAs are critical, since those might result in ignoring the relevant traffic light.
A false positive TL2LA results in regarding an additional traffic light.
This might result in stopping at an actual green light, which is less critical from a safety perspective.
We design our proposed methods to account for the adjusted Bayes risk.
In our pattern-based contribution method, we design $h$ in favor of a higher recall.
In our rejection method, we assume $\operatorname{assign}(s,l) = 1$ as the null hypothesis.
It is important to highlight that our proposed methods do not cover all edge cases and are not suitable for deployment (see Section \ref{sec:limitations} for limitations).

In addition to predicting the assignment, it is essential to assess the confidence of the output.
From an information-theoretical perspective, confidence is based on how many different combinations of traffic light states have been observed, and the consistency between traffic light states and motion patterns given the derived TL2LAs.
By the law of large numbers, the likelihood of seeing all state combinations rises with an increasing amount of samples and the accuracy runs into saturation.
Our rejection method uses a hypothesis test that provides a p-value as a function of the number of samples and the consistency.
Therefore, it can be used as a direct measure of confidence.

\section{Implementation on Dataset} \label{sec:implementation}
This section proposes the use of widely available motion prediction datasets for semantic map learning.
Motion prediction datasets are compared regarding their use for deriving the TL2LA.
We perform a dataset transformation on the Lyft Level 5 dataset~\cite{WovenPlanet2020} and explain data preparation steps.

\subsection{Motion Prediction Datasets for Semantic Map Learning} \label{sec:dataset}
\begin{figure}[tpb]
	\vspace{1.6mm}
	\includegraphics[width=1\columnwidth,trim={0 0 0 0.2cm},clip]{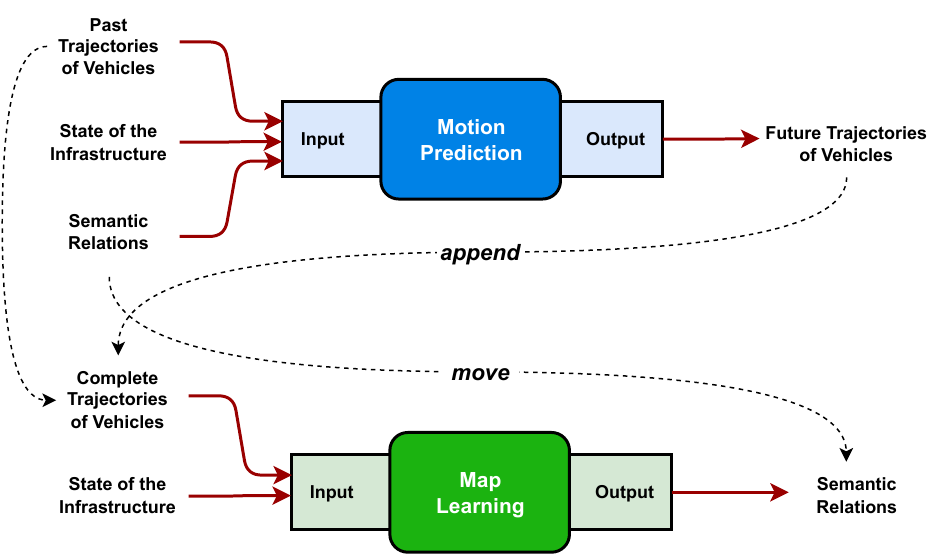}
	\caption{Visualization of the dataset transformation that converts motion prediction datasets into a representation suitable for
	learning map semantics from motion data. Semantic relations are used as outputs instead of inputs.}
	\label{fig:coherence_motion_froecasting_map_learning}
	\vspace{-1.6mm}
\end{figure}

The problem of motion prediction is defined by a function:
\begin{equation}
	\operatorname{pred} \left( V, S, L, \operatorname{pos}_{0:t}, \operatorname{state}, \operatorname{boundary}, \operatorname{assign} \right) = \operatorname{pos}_{t+1:T}
\end{equation}
It predicts future trajectories $\operatorname{pos}_{t+1:T}$ from past trajectories $\operatorname{pos}_{0:t}$.
Additional inputs often include lanes ($L$) with boundaries ($\operatorname{boundary}$), traffic lights $S$ with their states ($\operatorname{state}$), and the TL2LAs ($\operatorname{assign}$).
We show that specific datasets for motion prediction can be transformed into datasets for semantic map learning.
A comparison with function $g$ for the TL2LA problem (c.f., Equation \ref{eq:tl2la_dataset}) shows the required steps to transform the dataset (also visualized in Figure \ref{fig:coherence_motion_froecasting_map_learning}):
\begin{enumerate}
	\item Future trajectories are appended to past trajectories to form complete vehicle trajectories as input:\\
	 $\operatorname{pos} = \operatorname{pos}_{0:T} = \left[ \operatorname{pos}_{0:t},  \operatorname{pos}_{t+1:T} \right]$
	\item Semantic relations, specifically TL2LAs, are used as output instead of input: $Y = \operatorname{assign}$
\end{enumerate}

Datasets for motion prediction are usually annotated from recordings of a measurement vehicle (referred to as ego).
The partial functions $\operatorname{state}$ and $\operatorname{pos}$ are only defined for traffic lights and vehicles in the field of view of the measurement vehicle.
We can use this locality to formulate a condition for a pair of traffic light $s$ and lane $l$:
\begin{equation}
	\label{eq:condition}
	\exists t \in T: \operatorname{state}(s,t) \land \operatorname{pos}(v,t) \land \left( \operatorname{loc}(v,t) = l \right)
\end{equation}
It requires that there exists a point in time in the dataset where a state for $s$ and position for a vehicle $v$ are detected while $v$ is on $l$.
This condition is not met for most pairs in motion prediction datasets, since most pairs of traffic lights and lanes are not part of the same intersection.
Predicting a TL2LA is not useful in those cases where there is no evidence of motion patterns on that lane relative to the traffic light state.

\subsection{Comparison of available Datasets}
Table \ref{tab:comparison_features} shows which available motion prediction datasets exhibit which features.
Only the datasets from Waymo and Lyft allow us to derive TL2LA from motion data.

\begin{table}[tpb]
	\vspace{4mm}
	\centering
	\caption{Comparison of relevant dataset features}
	\newcolumntype{C}[1]{>{\centering\let\newline\\\arraybackslash\hspace{0pt}}m{#1}}
	\label{tab:comparison_features}
	\resizebox{\columnwidth}{!}{%
		\begin{tabular}{%
			>{\raggedright}p{.33\columnwidth}%
			C{.2\columnwidth}%
			C{.12\columnwidth}%
			C{.11\columnwidth}%
			C{.08\columnwidth}%
			}
			\toprule
			\textbf{Feature}                  & \textbf{Argoverse 2} & \textbf{nuScene} & \textbf{Waymo} & \textbf{Lyft} \\
			\midrule
			Lane geometry                     & \cmark                 & \cmark           & \cmark                & \cmark             \\

			Lane topology                     & \cmark                 & \cmark           & \cmark                & \cmark             \\

			Vehicle trajectory                  & \cmark                 & \cmark           & \cmark                & \cmark             \\

			Vehicle to lane mapping             & \xmark                 & \xmark           & \xmark                & \xmark             \\

			Traffic light state               & \xmark                 & \cmark           & \cmark                & \cmark             \\

			Traffic light geometry & \xmark                 & \xmark           & \xmark                & \cmark             \\

			TL2LA  & \xmark                 & \xmark           & \cmark                & \cmark             \\
			\bottomrule
		\end{tabular}
	}
\vspace{-1em}
\end{table}

To perform statistics on the motion patterns relative to a traffic light state, a suitable dataset needs to have many recordings of the same intersection.
The datasets in Table \ref{tab:comparison_metadata} do not provide information at the intersection level.
Instead, we provide a different metric by defining a spatial recording density~$\rho$ as follows:
\begin{equation}
	\rho = \frac{N \cdot \tau}{R}
\end{equation}
where~$N$ is the number of scenes, ~$\tau$ is the duration of a scene, and~$R$ is the length of unique roadways.
Table \ref{tab:comparison_metadata} shows the resulting values.
The Lyft dataset has many scenes on a small set of unique roadways.
Hence, the resulting spatial recording density is high at $\SI{405000}{\s\per\km}$.
Its annotated map contains 12 intersections with 22 annotated intersection branches and 279 TL2LAs in total.

\begin{figure}[tpb]
	\vspace{2mm}
	\centering
	\includegraphics[width=0.85\columnwidth]{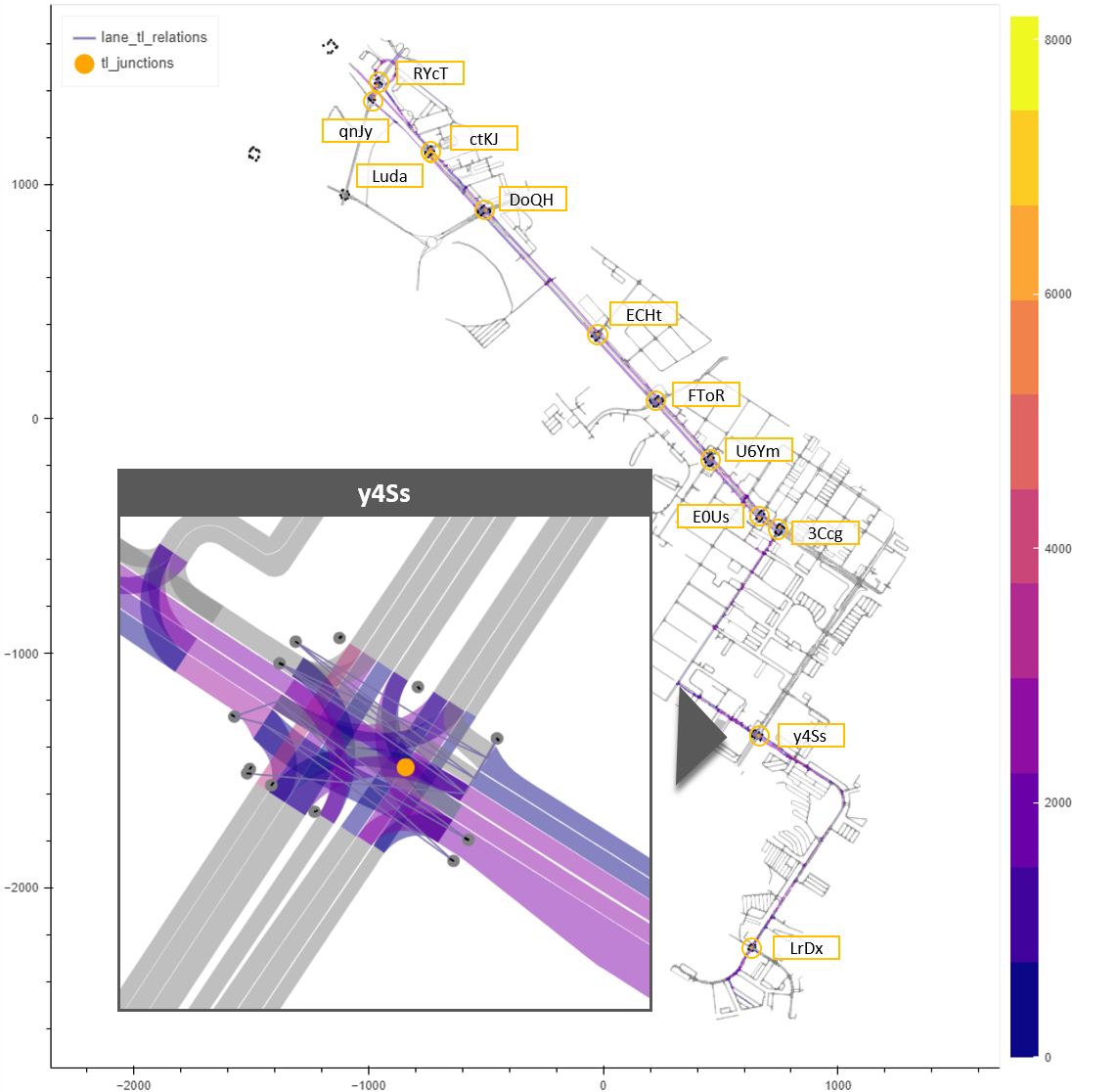}
	\caption{Visualization of the semantic map of the Lyft dataset, including lane segments, traffic light geometries and TL2LAs. The heatmap indicates how often the ego vehicle traveled a specific lane segment.}
	\label{fig:heatmap}
	\vspace{-2mm}
\end{figure}

Figure \ref{fig:heatmap} visualizes the semantic map of the Lyft dataset in Palo Alto, California.
The color represents the number of scenes recorded by a fleet of 20 vehicles.
The heatmap for the zoomed-in intersection \textit{y4Ss} shows that the ego vehicle does not travel all lanes. 
Therefore, using only the motion patterns of the ego vehicle will not give information for all TL2LAs in the semantic map.
In addition, the TL2LAs in the Lyft dataset are only partially annotated to cover intersection branches traveled by the ego vehicle.
Another geospatial requirement is that a suitable dataset should provide the map in a global coordinate system to avoid the need to localize the scenes to each other.
This is also one evaluation criterion listed in Table \ref{tab:comparison_metadata} and fulfilled by Argoverse 2, nuScenes and Lyft datasets.
In summary, the Lyft dataset best meets all criteria and is used in the further course of this work.

\subsection{Dataset Preparation}
Making use of the Lyft dataset for map learning requires some additional preprocessing steps.
Also, the vast amount of available scenes allows removing data points that provide no clear evidence, e.g., due to noise.
The following steps are performed:
\begin{itemize}
	\item Lanes are represented in the Lyft dataset as segments. We manually label lanes by forming sequences of lane segments from the intersection entry \SI{20}{\m} backward.
	This also filters out overlapping lane segments and asserts unique geometric assignments from vehicle to lane by $\operatorname{loc}$).
	\item Only vehicles on the same intersection branch and at a certain distance to the intersection are considered.
	      Those are close to the intersection entry, so their motion pattern is highly related to the traffic light state.
	\item The TL2LA label is only kept for those pairs of traffic lights and lanes that meet the condition stated in Equation \ref{eq:condition}.
	      This way, pairs not part of the same intersection are filtered out and 279 pairs remain in the Lyft dataset.
\end{itemize}

\section{Experiments} \label{sec:experiments}
We evaluate our methods on two subsets of the Lyft dataset, considering 1.) only trajectories of the recording vehicle, referred to as ego vehicle and 2.) all vehicle trajectories.
Considering only ego trajectories yields reliable motion patterns and avoids tracking and occlusion issues.
However, this limits the predictable set of TL2LAs, since not all lanes of each intersection were traveled by the ego vehicle in the Lyft dataset.

\subsection{Metrics}\label{sec:metrics}
Accuracy (Acc), precision (Prec), recall and F-Score (F\textsubscript{1}) are evaluated. 
Recall is most important, because false negative TL2LAs are most critical for safety (see Section \ref{sec:safety}).
Precision is still essential, since false positive TL2LAs might yield unnecessary stops at actual green light.

\subsection{Quantitative Results}\label{sec:quantitative_results}
\begin{table*}[b]
	\centering
	\caption{Quantitative Results}
	\label{tab:quantitative_results}
	\newcolumntype{C}[1]{>{\centering\let\newline\\\arraybackslash\hspace{0pt}}m{#1}}
	\begin{tabular}{llccccccc}
		\toprule
		\textbf{Method}               & \textbf{Scope} & \textbf{Scenes} & \textbf{Vehicles} & \textbf{TL-Lane Pairs} & \textbf{Acc[\%]} & \textbf{Prec[\%]} & \textbf{Recall[\%]} & $\mathbf{F_1}$[\%] \\
		\midrule
		\multirow{2}{*}{Naive Baseline} & ego only        & 109k            & 109k              & 55                     & 81.8             & 81.8              & \textbf{100}        & 90.0               \\
		                                & all vehicles    & 109k            & 109k              & 271                    & 66.4             & 66.4              & \textbf{100}        & 79.8               \\
		\midrule
		\multirow{2}{*}{\makecell[l]{Pattern-based                                                                                                                                                         \\
		Contribution}}                  & ego only        & 64k             & 64k               & 55                     & 83.6             & 84.6              & 97.8                & 90.7               \\
		                                & all vehicles    & 90k             & 10M               & 271                    & 76.4             & \textbf{83.2}     & 80.4                & 81.8               \\
		\midrule
		\multirow{2}{*}{Rejection}  & ego only        & 13k             & 13k               & 55                     & \textbf{85.5}    & \textbf{84.9}     & \textbf{100}        & \textbf{91.8}      \\
		                                & all vehicles    & 42k             & 124k              & 271                    & \textbf{80.8}    & 78.3              & 98.3                & \textbf{87.2}      \\

		\bottomrule                                                                                                                                                                                        \\
	\end{tabular}
\end{table*}
Table \ref{tab:quantitative_results} reports the quantitative results of the three methods on the Lyft dataset.
The rejection method can only classify 271 out of the 279 pairs of traffic lights and lanes due to a limitation of the Lyft dataset.
All methods were evaluated on that set of 271 pairs to have a direct comparison.
\subsubsection{Naive Baseline Method}
The Lyft dataset consists of 109k scenes at traffic light-controlled intersections that include detections of traffic light states.
The naive baseline method can use all 109k scenes and returns the most probable TL2LA class in the Lyft dataset, which is $\operatorname{assign}(s,l) = 1$ for every pair of traffic light and lane.
Therefore, it reaches 100~\% recall, as well as 90.0~\% F\textsubscript{1} score on ego trajectories only and 79.8~\% F\textsubscript{1} score on all vehicles.

\subsubsection{Pattern-based Contribution Method}
The pattern-based contribution method only uses scenes within a certain distance to the intersection.
From the 109k scenes, 90k are available when considering all vehicles and 64k are available when considering only the ego vehicle.

The pattern-based contribution method roughly matches the performance of the naive baseline method on ego trajectories only, but outperforms it on all vehicles with an 81.8~\% vs. 79.8~\% F\textsubscript{1} score.
This indicates that the pattern-based contribution can exploit evidence from the motion patterns of vehicles to verify or falsify a TL2LA.
The comparison with our rejection method shows a lower performance of the pattern-based contribution method considering their accuracy, recall, and F\textsubscript{1} scores.
Only its precision is considerably higher with 83.2~\% vs. the precision of the rejection method with 78.3~\%.
We believe that a more sophisticated approach is needed than aggregating predictions of single scenes by the predicted class majority.

\begin{figure}[tpb]
	\vspace{2mm}
	\centering
	\includegraphics[width=1\linewidth,trim={0 0.22cm 0 0.22cm},clip]{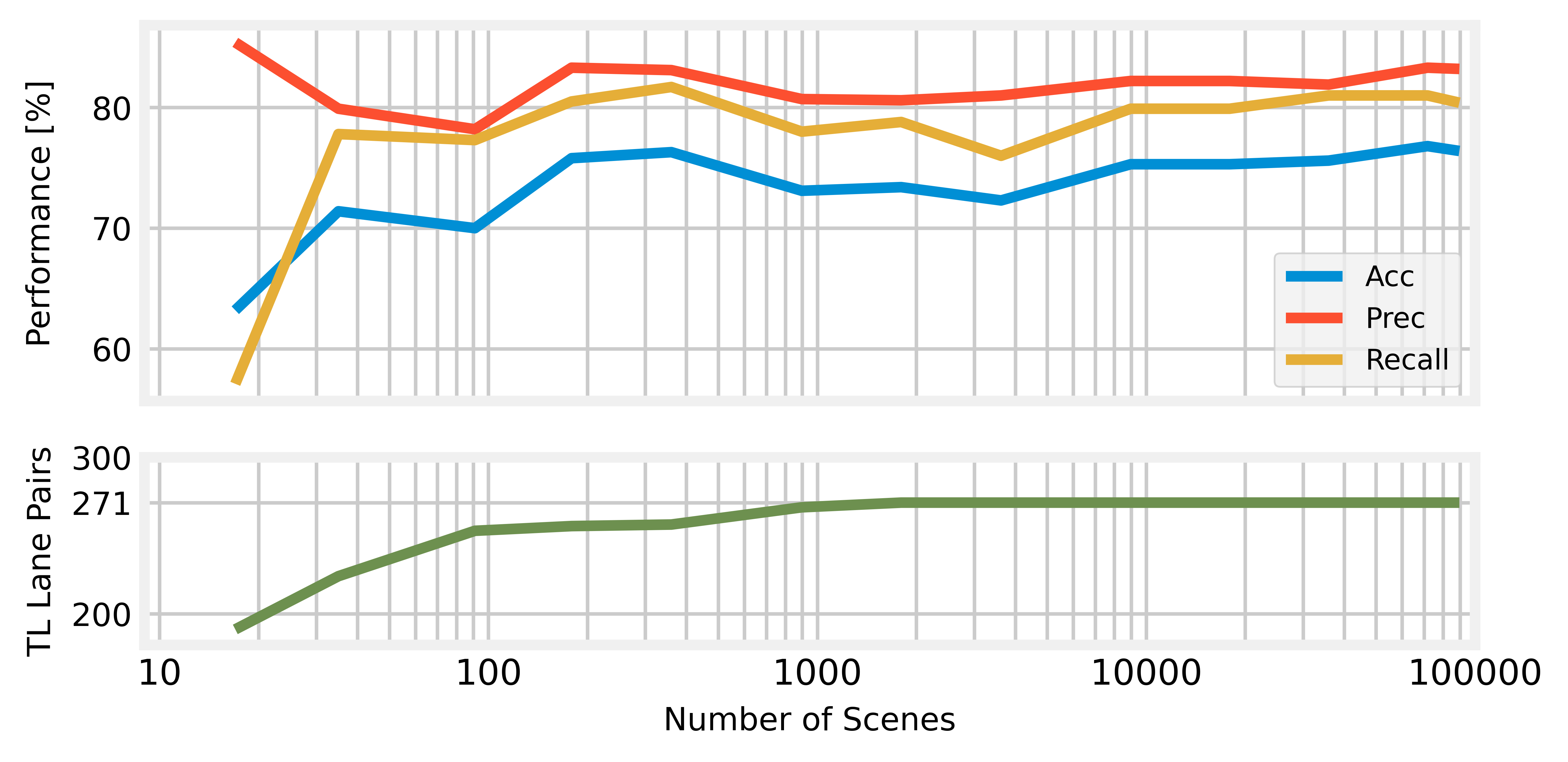}
	\caption{Visualization of the performance over the number of analyzed scenes of the pattern-based contribution method.}
	\label{fig:heristic_performance}
	\vspace{-2mm}
\end{figure}

Furthermore, we investigate the effect of the number of analyzed scenes that was theoretically described in Section \ref{sec:safety}.
Figure \ref{fig:heristic_performance} visualizes model performance as a function of the number of analyzed scenes of the pattern-based contribution considering all vehicles.
For the initial 1000 scenes, only a subset of the pairs of traffic lights and lanes can be predicted. 
Thus, the metrics cannot directly be compared but indicate a trend.
Accuracy, precision and recall reach saturation with an increasing number of scenes.

\subsubsection{Rejection Method}
The rejection method extracts information from scenes where vehicles cross an intersection while ego detects a traffic light state.
13k scenes can be used for the rejection method when considering only ego trajectories.
The reason is that only data points with a distance $< 1$ meter to the intersection entry are used, where not all traffic lights are in the field of view of the Lyft vehicles.
This is primarily not a limitation of our method but of the Lyft sensor set.
Hence, the rejection method on ego trajectories reaches a perfect recall of 100~\%, but can only classify 55 traffic light and lane pairs.

When using all vehicle trajectories, ego can record other vehicles driving through an intersection from a distance while detecting the traffic light states.
This yields 42k usable scenes, and 271 out of 279 traffic light and lane pairs can be classified.
The rejection method outperforms the pattern-based contribution method with an F\textsubscript{1} score of 87.2~\% and a close to perfect recall of 98.3~\%.
The precision is 78.3~\% because the null hypothesis of a true TL2LA is only rejected with a significant amount of red-light passes.
Although precision is lower compared to the pattern-based contribution method, the rejection method is preferable given the importance of a high recall.

\subsection{Qualitative Results}\label{sec:qualitative_results}
Our approach revealed 20 incorrect TL2LA labels in the Lyft dataset.
Assignments are missed in few instances and in scene 474 a dedicated left turn lane is incorrectly assigned to the traffic light controlling the straight direction.
The ego vehicle is located on the northwest branch and can detect the state of the traffic light \textit{icM8}, which belongs to the southwest branch.
While the semantic map assigns the traffic light \textit{icM8} to the oncoming lanes southwest of the intersection, there must be an assignment to the left-turn lane \textit{H+dt} of the northwest intersection branch.
Accordingly, our approach predicts  $\operatorname{assign}(\textit{icM8}, \textit{H+dt}) = 1$, whereas the label defines $\operatorname{assign}(\textit{icM8}, \textit{H+dt}) = 0$
This is a false negative error in the label.
False negative errors are critical because a relevant traffic light is not obeyed.
In this example, the vehicle would perform a left turn even if traffic light \textit{icM8} was red.
Since the geometry of the traffic light \textit{UJ52} cannot be differentiated in the two-dimensional space from the geometry of \textit{icM8}, it seems obvious that the wrong traffic light was selected during the manual labeling process of the semantic map.
This shows that human annotations are error-prone even for small coverages.
We manually corrected the false labels and performed all experiments on the corrected dataset.

\begin{figure}[tpb]
	\vspace{2mm}
	\includegraphics[width=1\columnwidth]{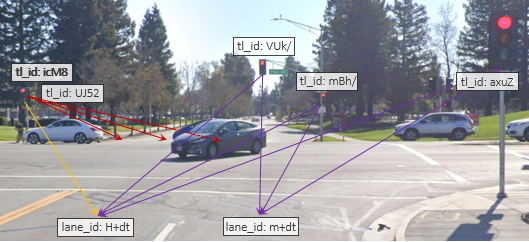}
	\caption{Lyft dataset error in scene 474 with the true positive (purple), false negative (yellow) and false positive (red) TL2LA labels \cite{GoogleStreetView}.}
	\label{fig:invalid_tl2la_streetview}
	\vspace{-2mm}
\end{figure}

Figure \ref{fig:invalid_tl2la_streetview} shows the real-world perspective (northwest branch) of the ego vehicle in scene 474.
Additionally, the TL2LAs are visualized. 
Yellow shows the assignment of traffic light \textit{icM8} to lane \textit{H+dt}, which is not labeled in the semantic map (false negatives), and red indicates erroneously labeled TL2LAs (false positives).

\begin{figure}[tpb]
	\raggedleft
	\vspace{4mm}
	\includegraphics[width=0.9\linewidth]{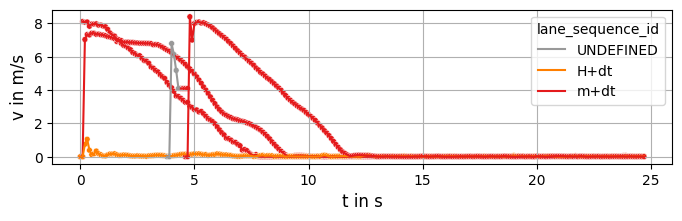}\\
	\vspace{-1em}
	\raggedleft
	\includegraphics[width=0.98\linewidth]{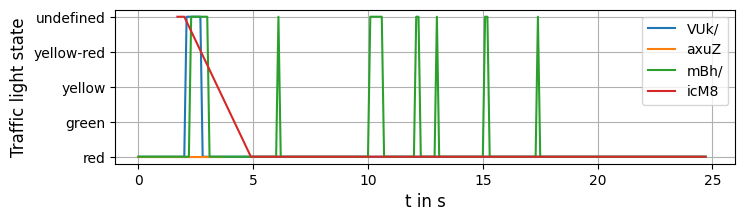}\\
	\caption{Graph of vehicle velocities and traffic light states over the time for scene 474. Based on motion patterns and detected traffic light states, the pattern-based contribution method assumes TL2LAs between the lanes \textit{m+dt}, \textit{H+dt} and their traffic lights \textit{VUk/, axuZ, mBh/, icM8}.}
	\label{fig:v_t_diagram}
	\vspace{-2mm}
\end{figure}

Figure \ref{fig:v_t_diagram} shows the velocity-time and the traffic light state-time diagrams of scene 474.
The colors in the velocity-time diagram represent the assignment of the detected vehicles to the two lane sequences in front of the intersection.
Grey indicates that the position of a vehicle could not be mapped to a lane sequence.
The state-time diagram visualizes the state of the detected traffic lights.
Due to occlusion, the traffic light states are noisy and cannot be detected consistently.

\section{Limitations} \label{sec:limitations}
Our approach to deriving the TL2LA from motion data has certain limitations described in the following.

\subsubsection{Availability of Traffic Light States and Motion Patterns}
A TL2LA is only derived between lanes that vehicles have been recorded on and traffic lights whose state has been detected.
We assume that a sensor set is chosen that can detect all relevant traffic lights.
Eventually, pairs of traffic lights and lanes that are not covered in the dataset can be conservatively assumed to have a TL2LA.
This way, all detected traffic lights are considered, and the safety-critical case of disregarding a true TL2LA is avoided.

\subsubsection{Disambiguation of synchronized Traffic Lights}
No disambiguation is possible if two traffic lights are only recorded in the same state, even if their lane assignments differ.
This can be resolved by combining the motion data-based approach with a geometry-based and inlay-based approach.

\subsubsection{Traffic Light-independent Rules}
There are other corner cases such as flashing red lights, which are treated as an all-way stop, vehicles running a red light, or police controlling traffic.
Examples are scenes 495, 597, and 1719 in the Lyft dataset, where all traffic lights for the straight lanes are detected as red, but vehicles on all lanes pass the intersection.
The applied methods need to be robust to handle such outliers.

\section{Conclusion} \label{sec:conclusion}
In this paper, we presented a novel solution to the problem of learning traffic light to lane assignments by using motion data.
Both a pattern-based contribution and a rejection method were implemented and validated to show the trade-off between precision and recall.
We found that the rejection method is very effective regarding safety considerations.
For future work, a more sophisticated approach could use a generic graph encoding \cite{Monninger2023} and formulate the TL2LA task as link prediction in graphs.

Using motion patterns and traffic light states, our proposed approach derives the TL2LA independent of the geometric constellation.
In deployment, combining this motion-based approach with a geometric approach can yield the best results by either using both sources of information in an integrated model or by having redundancy.
Additionally, we proposed a dataset transformation to enable the use of available motion prediction datasets for this task.
By providing an API for the Lyft Level 5 dataset, we encourage the research community to invent robust approaches that meet the desired safety requirements.

\pagebreak

\bibliographystyle{IEEEtran}
\bibliography{Literature}

\end{document}